\documentclass[runningheads]{llncs}

\usepackage{multirow}
\usepackage{floatrow}
\usepackage{blindtext}

\usepackage{graphicx}

\usepackage{amsmath,amssymb} 
\usepackage{color}
\usepackage[utf8]{inputenc}

\begin{document}
\newfloatcommand{capbtabbox}{table}[][\FBwidth]

\title{CentralNet: a Multilayer Approach for Multimodal Fusion}
\titlerunning{CentralNet}
\authorrunning{V. Vielzeuf, A. Lechervy, S. Pateux, F. Jurie}

\author{Valentin Vielzeuf\inst{1,2} \and
Alexis Lechervy\inst{2} \and
Stéphane Pateux\inst{1} \and
Frédéric Jurie\inst{2}}
\institute{Orange Labs \and Normandie Univ., UNICAEN, ENSICAEN, CNRS\\
	\email{ \{valentin.vielzeuf,stephane.pateux\}@orange.com}\\
	\email{ \{alexis.lechervy,frederic.jurie\}@unicaen.fr}
}

\maketitle

\begin{abstract}
This paper proposes a novel multimodal fusion approach, aiming to produce best possible decisions by integrating information coming from multiple media. While most of the past multimodal approaches either work  by projecting  the features of different modalities into the same space, or by coordinating the representations of each modality through the use of constraints, our approach borrows from both visions. More specifically, assuming each modality can be processed by a separated deep convolutional network, allowing to take decisions independently from each modality, we introduce a central network linking the modality specific networks. This central network not only provides a common feature embedding but also regularizes the modality specific networks through the use of multi-task learning. The proposed approach is validated on 4 different computer vision tasks on which it consistently improves the accuracy of existing multimodal fusion approaches.
\end{abstract}

\section{Introduction and Related Work}

Multimodal approaches are key elements for many computer vision applications, from video analysis to medical imaging, through natural language processing and image analysis. The main motivation for such approaches is to extract and combine relevant information from the different modalities and hence take better decisions than using only one. The recent literature abounds with examples in different domains such as video classification~\cite{abu2016youtube,wang2017truly}, emotion recognition~\cite{dhall2012collecting,ringeval2017summary,hu2017learning}, human activity recognition~\cite{simonyan2014two}, or more recently food classification from pictures and recipes~\cite{kiela2018efficient}.

The literature on multimodal fusion~\cite{atrey2010multimodal,andrew2013deep,lahat2015multimodal} usually distinguishes the methods accordingly with the level at which the fusion is done (typically early vs late fusion). There is no consensus on which level is the best, as it is task dependent. For instance, Simonyan~\textit{et al.}~\cite{simonyan2014two} propose a two stream convolutional neural network for human activity recognition, fusing the modalities at prediction level. Similarly, for audiovisual emotion recognition, several authors report better performance with late fusion approaches \cite{kim2017multi,vielzeuf2017temporal}. In contrast, Arevalo~\textit{et al.}~\cite{arevalo2017gated} propose an original Gated Multimodal Unit to weight the modalities depending on the input and achieve state of the art results on a textual-visual dataset, while Chen~\textit{et al.}~\cite{chen2017multimodal} follow an early fusion hard-gated approach for textual-visual sentiment analysis. 

Opposing early and late fusion is certainly too limited a view on the problem. As an illustration, Neverova~\textit{et al.}~\cite{neverova2014multi} applies a heuristic consisting in fusing similar modalities earlier than the others.
Several hybrid or multilayer approaches have also been proposed, such as the approach of Yang~\textit{et al.}~~\cite{yang2016multilayer} doing fusion by boosting across all layers on human activity videos.  Cătălina Cangea~\textit{et al.}~\cite{xflow} propose a multilayer cross connection from 2D to 1D to share information between modalities of different dimensions. A multilayer method is also applied on text and image multimodal datasets in the paper of Gu~\textit{et al.}~\cite{gu2017learning}. Kang~\textit{et al.}~\cite{kang2017contextual} use a multilayer approach, aggregating several layers of representation into a contextual representation. These hybrid methods can be viewed as learning a {\em joint representation}, following the classification made by Baltruvsaitis~\textit{et al.}~\cite{baltruvsaitis2018multimodal}. With this type of approach the different modalities are projected  into the same multimodal space, e.g. using concatenation, element-wise products, etc.

Baltruvsaitis~\textit{et al.}~\cite{baltruvsaitis2018multimodal} oppose {\em joint representations} with {\em coordinated representations} where some constraints between the modalities force the representations to be more complementary. These constraints can aim at maximizing the correlation between the multimodal representations, as in Andrew~\textit{et al.}~ \cite{andrew2013deep} who propose a deep Canonical Correlation Analysis method. On their side, Chandar~\textit{et al.}~ \cite{chandar2016correlational} propose CorrNet using autoencoders. Neverova~\textit{et al.}~ \cite{neverova2016moddrop,li2016modout} propose modDrop and modout regularization, consisting in dropping modalities during the training phase. Finally, Hu~\textit{et al.}~ \cite{hu2017learning} applies an ensemble-like method to solve the problem of multimodal fusion for emotion classification.

This paper borrows from both visions, namely the joint representations and the coordinated representations. Our fusion method builds on existing deep convolutional neural networks designed to process each modality independently. We suggest to connect these networks using an additional central network dedicated to the projection of the features coming from different modalities into the same common space. In addition, the global loss allows to back propagate some global constraints on each modality, coordinating their representations. As an interesting property, the proposed approach automatically identifies which are the best levels for fusing the information and how these levels should be combined. The approach is multitask in the sense that it simultaneously tries to satisfy per modality losses as well as the global loss defined on the joint space. 

The rest of the paper is organized as follows: the next section presents our contribution while Section~\ref{Experiments} gives an experimental validation of the approach.

\begin{figure}[tb]
\centering
\includegraphics[width=\linewidth]{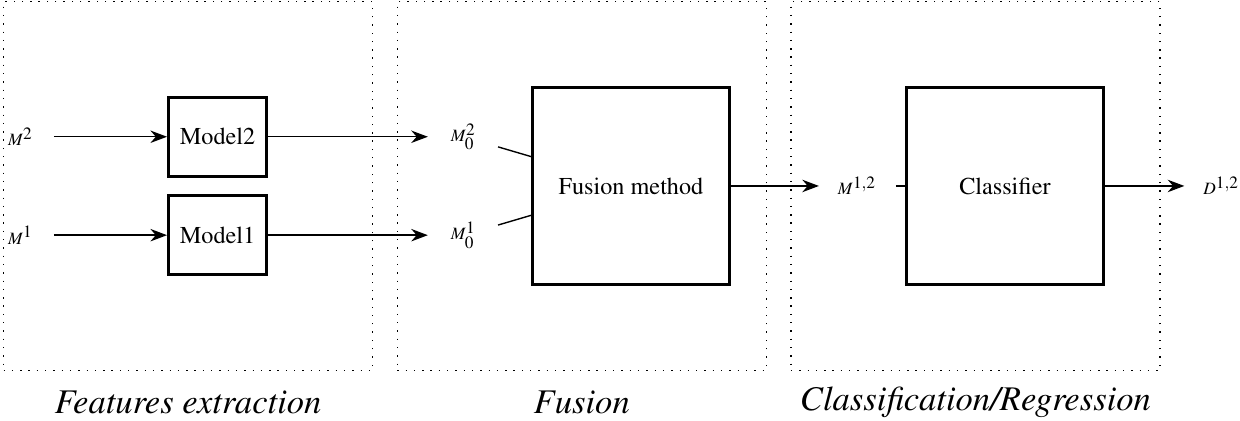}\\
\caption{\textbf{Generic representation of a multimodal fusion model.} $M^1$ and $M^2$ respectively denote modality 1 and modality 2,  $M_0^1$ and $M_0^2$ are the modality features fed to the fusion method, $M^{1,2}$ is the joint representation produced by the fusion method, and $D^{1,2}$ the decision obtained from the joint representation.}
\label{resume}
\end{figure}

\section{CentralNet}
\label{Methods}
We refer to \textit{multimodal fusion} as the combination of information provided by different media, under the form of their associated features or the intermediate decisions. More formally, if  $M^1$ and $M^2$ denote the two media and $D^1$ and $D^2$ the decisions inferred respectively by $M^1$ and $M^2$, the goal is to make a better prediction $D^{1,2}$ using both $M^1$ and $M^2$. More than 2 modalities can be used. This paper addresses the case of classification tasks, but any other task, \textit{e.g.} regression, can be addressed in the same way.  

This paper focuses on the case of neural nets, for which the data are sequentially processed by a succession of layers. We assume having one neural net per modality, capable of inferring a decision from each  modality taken in isolation, and want to combine them. One recurrent question with multimodal fusion is where the fusion has to be done: close to the data (early fusion), at the decision level (late fusion) or in between. In  case of neural networks, the fusion can be done at any level between the input and the output of the different unimodal networks. For the sake of presentation, let us consider that the neural networks are split into 3 parts: the layers before the fusion (considered as being the feature generation part of the networks), the layers used for the fusion and finally the classification parts of the networks. This is illustrated by Figure~\ref{resume}. 

For simplicity, we assume that the extracted features (at the input of the fusion layers) have the same dimensionality. If it is not the case, the features can be projected, \textit{e.g.} with 1x1 convolutional layers or zero padded to give them the same size. In practice, the last convolution layers or the first dense layers of separately trained unimodal networks can be used as features.  

\begin{figure}[tb]
\begin{minipage}{\textwidth}
\centering
\includegraphics[width=0.98\linewidth]{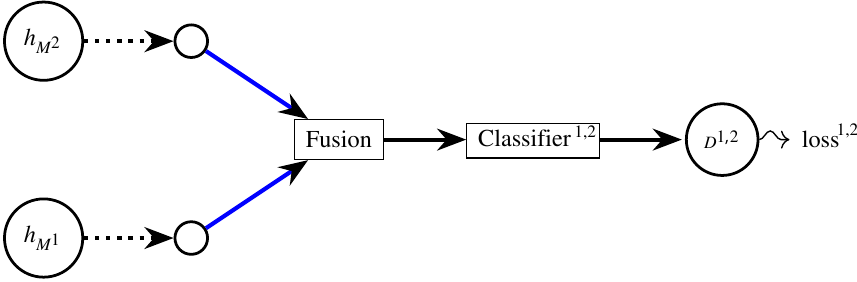}\\
(a)
\end{minipage}
\begin{minipage}{\textwidth}
\centering
\includegraphics[width=0.98\linewidth]{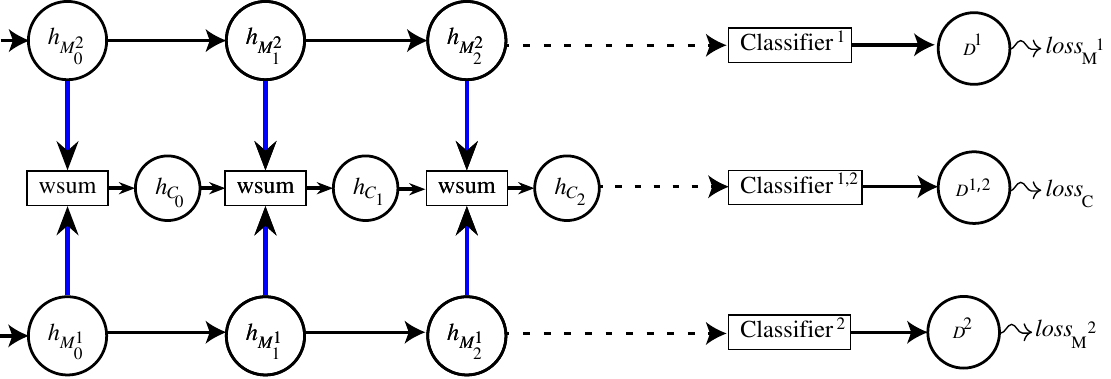}\\
(b)
\end{minipage}
\caption{\textbf{(a) Basic fusion method}, fusing the hidden representations of the modalities at a given layer and then using only joint representation. Fusing at a low-level layer is called early fusion while fusing at the last layer is called late fusion.
\textbf{(b) Our CentralNet fusion model}, using both unimodal hidden representations and a central joint representation at each layer. The fusion of the unimodal representations is done here using a learned weighted sum.
For the sake of simplicity, only the overall synoptic views of the architectures are represented. More details are provided in Section~\ref{Methods}.
}
\label{schemaCentral}
\end{figure}

\subsection{CentralNet Architecture}
The CentralNet architecture is a neural network which combines the features issued from different modalities, by taking, as input of each one of its layers, a weighted sum of the layers of the corresponding unimodal networks and of its own previous layers. This is illustrated in Figure~\ref{schemaCentral}(b). Such fusion layers can be defined by the following equation:
\begin{equation} h_{C_{i+1}} =  \alpha_{C_{i}} h_{C_{i}} + \sum_{k=1}^{n} \alpha_{M_{i}^{k}} h_{M_{i}^{k}} 
\label{mainEqu}
\end{equation}
where $n$ is the number of modalities, $\alpha$ are scalar trainable weights, $h_{M_{i}^{k}}$ is the hidden representation of each modality at layer $i$, and $h_{C_{i}}$ is the central hidden representation.
The resulting representation $h_{C_{i+1}}$ is then fed to an operating layer cell (which can be a convolutional or a dense layer followed by an activation function).  

Regarding the first layer of the central network ($i=0$), as we do not have any previous central hidden representation, we only weight and sum the representations of $M^1$ and $M^2$, issued from unimodal networks. 
At the output level, the last weighted sum is done between the unimodal predictions and the central prediction. Then, the output of the central net (classification layer) is used as the final prediction.

\subsection{Learning the CentralNet model}
All trainable weights of the unimodal networks, the ones of the CentralNet and the fusion parameters $\alpha_{M_{i}^{k}}$, are optimized together by applying a stochastic gradient descent using the Adam approach.
The global loss is defined as: 

\begin{equation} loss = loss_{C} +  \sum_{k} \beta_k loss_{M^{k}} \end{equation}

where $loss_{C}$ is the (classification) loss computed from the output of the central model and $loss_{M^{k}}$ the (classification) loss when using only modality $k$. The weights $\beta_k$ are cross validated (in practice, $\beta_k=1$ in all of our experiments). 

As already observed by Neverova~\textit{et al.}~ \cite{neverova2016moddrop}, when dealing with multimodal fusion it is crucial to maintain the performance of the unimodal neural networks. It is the reason why the global loss includes the unimodal losses. It helps generalizations by acting as a multitask regularization. We name this method "Multi-Task" in the rest of the paper. 

\subsection{Implementation details}
The $\alpha_{C_{i}}$ weights are initialized following a uniform probability distribution. Before training, the weighted sum is therefore equivalent to a simple average.

During our experiments, we also found out that rewriting  Eq.~(\ref{mainEqu}) as: 
\begin{equation}
h_{C_{i+1}} =  \alpha_{C_{i}} h_{C_{i}} + \alpha_{modalities}\sum_{k=1}^{n} \alpha_{M_{i}^{k}} h_{M_{i}^{k}}\end{equation} 
leads to better and stable performance. 

On overall, CentralNet is easy to implement and can build on the top of existing architectures already known to be efficient for each modality. The number of trainable parameters dedicated to the fusion is less important than in other previous multilayer attempts such as~\cite{gu2017learning}, which may help to prevent over-fitting. And even if the weighted sum is a simple linear operation, the network has the ability to learn complex joint representation, because of the non-linearity introduced by the central network.

Finally, the resulting values of the $\alpha$ allow some interesting interpretations on where the modalities are combined. For instance, getting $\alpha_{M_{i}^{k}}$ values close to 0 for $k > 0$ is equivalent to early fusion, while having all the $ \alpha_{C_{i}}$ close to 0 up to the last weighted sum would be equivalent to late fusion.

\section{Experiments}
\label{Experiments}
The proposed method is experimentally validated on 4 different multimodal datasets, namely Multimodal MNIST (a toy dataset), Audiovisual MNIST, Montalbano~\cite{escalera2014chalearn} and MM-IMDb~\cite{arevalo2017gated}, receiving each a separate section in the following. Each dataset is processed with a dedicated features extractor on which we plug our fusion method. Regarding the fusion networks, they are made of convolution+pooling or dense layers, with ReLU and batch normalization. The kernel size of the convolution is always 5x5 and the pooling stride is 2. 

The performance of the proposed method is compared to 5 different fusion approaches, ranging from the simplest baseline to recent state-of-the-art approaches. (a) 'Weighted mean' is the weighted average of single modality scores. The weights are considered as some parameters of the model, learnt with the rest of the model. (b) 'Concat' consists in concatenating the unimodal scores and inferring the final score with a single-layer linear perceptron. (c)  'Concat+Multi-Task' is the same as 'Concat' but uses the same Multi-Task loss as with the CentralNet. (d) 'Moddrop' is implemented following Neverova \textit{et al.}~\cite{neverova2016moddrop}. 
(e) The 'Gated Multimodal Unit' (GMU) is implemented following Arevalo \textit{et al.}~\cite{arevalo2017gated}.

In the following, to assess the statistical significance of our results, the performance is averaged over 64 runs. The confidence interval at 99\% is computed using the estimate of the standard deviation and the Student's law.
\begin{figure}[t]
    \centering
    \includegraphics[width=\linewidth]{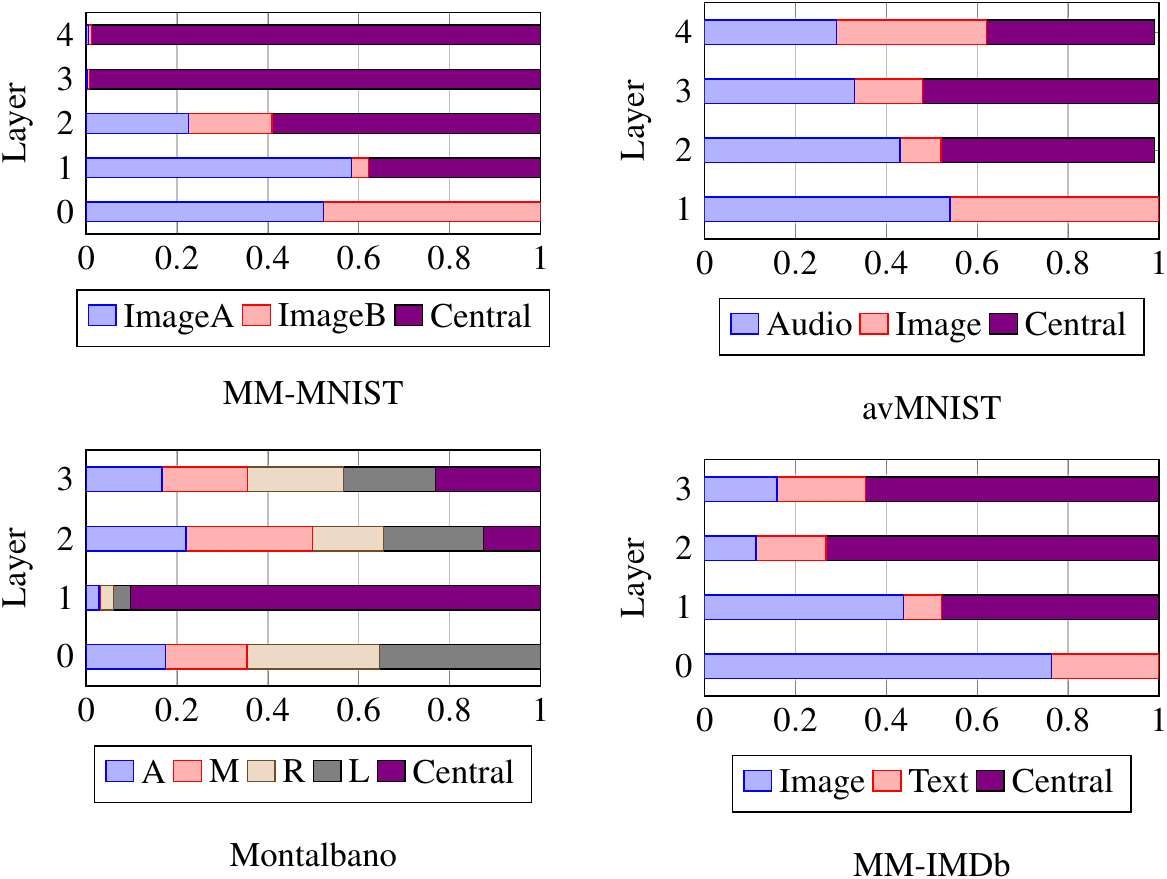}
    \caption{Visualization of the $\alpha_i$ weights after training. They are displayed as the percentage given to each modality and to the central hidden representations across layers and datasets. We observe that the learned fusion strategy is different for each dataset.}
    \label{all_weights}
\end{figure}

\subsection{Multimodal MNIST}
\begin{figure}[tb]
\begin{floatrow}
\ffigbox[.38\textwidth]{%
  \includegraphics[scale=0.5]{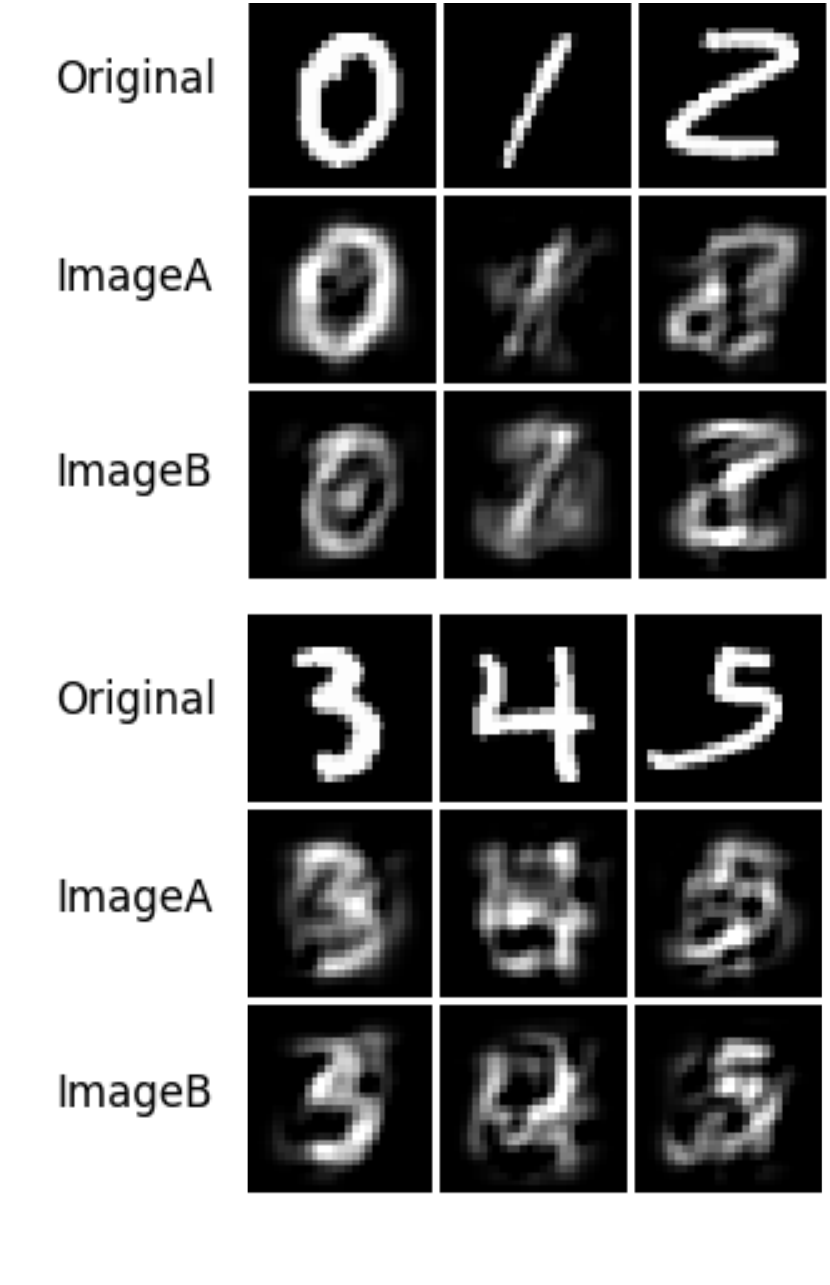}
}{%
  \caption{MM-MNIST: some examples generated with half of the energy per modality and no sharing.}%
  \label{generatedModas}
}
\capbtabbox{%
\centering
\begin{tabular}{lll}
\hline
Method & Errors      & \begin{tabular}[c]{@{}l@{}}Fusion\\ Layer\end{tabular}      \\ \hline
Baseline                      & 66 $\pm 1.5$ & \_ \\ 
Ensemble 2 classifiers        & 64 $\pm 1.3$ & \_ \\ 
Ensemble 3 classifiers        & 60 $\pm 1.0$ & \_ \\ \hline
\textbf{Fusion} \textbf{Subtract}      & \textbf{64 $\mathbf{\pm 1.8}$} & \textbf{2}  \\
Fusion Sum                    & 68 $\pm 2.1$         & 2           \\
Fusion Prod                   & 71 $\pm 2.1$         & 2           \\ \hline
Fusion+Ensemble  Subtract                & 63 $\pm 1.5$          & 1           \\
\textbf{Fusion+Ensemble} \textbf{Sum}           & \textbf{56 $\mathbf{\pm 2.1}$} & \textbf{0}  \\
Fusion+Ensemble  Prod                    & 63 $\pm 1.5$          & 2           \\ \hline
\hline
Baseline on one modality                 & 230  $\pm 2.7$        & \_          \\
Concat + Multi-Task                      & 62 $\pm 1.2$          & \_          \\
Moddrop\cite{neverova2016moddrop}         & 60 $\pm 1.5$          & \_          \\
Gated Multimodal Unit\cite{arevalo2017gated}            & 68 $\pm 1.8$          & \_          \\ 
\textbf{CentralNet} &                     \textbf{53 $\mathbf{\pm 1.2}$}             & \textbf{\_} \\ \hline

\end{tabular}

}{%
  \caption{Number of errors on the MM-MNIST test set for different methods, using 50\% energy per modality and 50\% of shared vectors.}
  \label{mMNIST_results}
}
\end{floatrow}
\end{figure}

The 'Multimodal MNIST' dataset is a toy dataset made of pairs of images (A,B), computed from the MNIST dataset. A and B are supposed to be 2 views of the same MNIST image but from different (artificially generated) modalities. We produce them by computing a Principal Component Analysis of the original MNIST dataset. Each one of the 2 (artificial) modalities is created by associating with it a set of singular vectors. This allows to control the amount of energy provided to each modality, which is the sum of the \textit{energy} contained in the chosen vectors, and the \textit{share ratio}, defined here as the percentage of the singular vectors shared between modalities. Figure~\ref{generatedModas} shows some of these generated image pairs. The original MNIST contains 55000 training samples and 10000 test samples. We transformed all of these images into pairs of 28x28 images, following the process explained above.

Several authors, \textit{e.g.},~\cite{andrew2013deep,neverova2016moddrop,chandar2016correlational,li2016modout}, generate a multimodal version of MNIST by dividing MNIST images into several smaller images (typically quarter of images) which are each considered as modalities. In contrast, our approach has the advantage of allowing to control two important factors: the amount of information per modality and the dependence between modalities. 

\begin{figure}[tb]
    \centering
    \includegraphics[width=\linewidth]{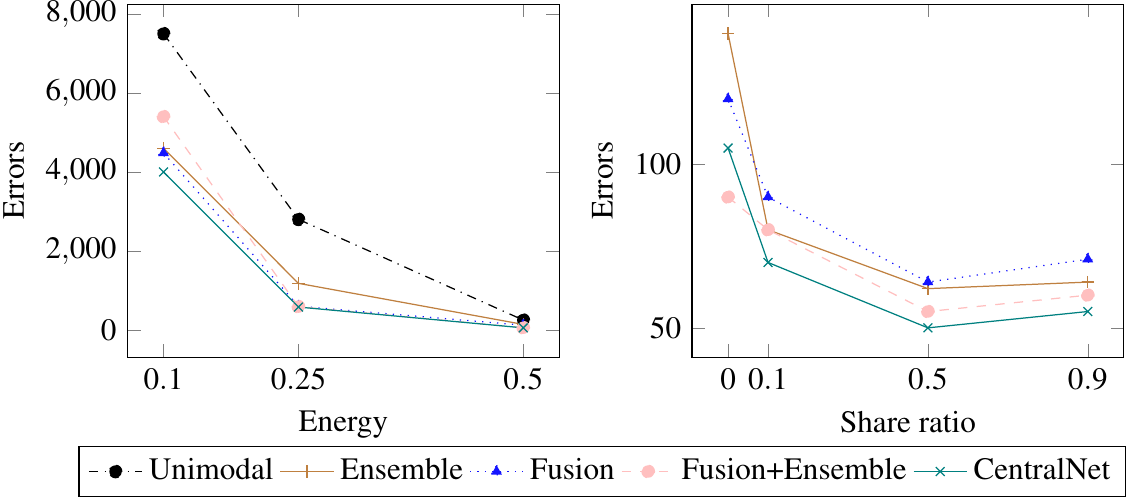}
    \caption{Errors as a function of the energy per modality (left-hand side, share ratio=0.5) and of the share ratio (right-hand side, energy=0.5), for different fusion methods. Better viewed in color.}
    \label{energy}
\end{figure}

\begin{table}[tb]
\centering
\begin{tabular}{ll|l|l|l|l}
\hline
\multicolumn{2}{l|}{Image A}           & \multicolumn{2}{l|}{Central} & \multicolumn{2}{l}{Image B} \\ \hline
\multicolumn{1}{l|}{Type}  & Size     & Type        & Size           & Type        & Size           \\ \hline
\multicolumn{1}{l|}{Conv}  & 14x14x32 & Conv        & 14x14x32       & Conv        & 14x14x32       \\
\multicolumn{1}{l|}{Conv}  & 7x7x64   & Conv        & 7x7x64         & Conv        & 7x7x64         \\
\multicolumn{1}{l|}{Dense} & 1024     & Dense       & 1024           & Dense       & 1024           \\
\multicolumn{1}{l|}{Pred}  & 10       & Pred        & 10             & Pred        & 10             \\ \hline
\end{tabular}
\caption{The architecture of the CentralNet for the mMNIST dataset."Dense" layers are fully-connected layers followed by a ReLU activation, while "Pred" layers are fully-connected layers followed by softmax activation.}

\label{mMNIST_archi}
\end{table}

The unimodal neural network architecture used with this dataset is the LeNet5 neural network~\cite{lecun1998gradient}. It achieves 95 errors on the MNIST test set~\cite{lecun1998gradient}. The architecture is composed of two convolutional layers, followed by two fully connected layers. In our version, batch normalization and dropout are added to further improve its performance.
We measure the performance by counting how many of the 10000 images of the MNIST test set are misclassified.
The CentralNet architecture in this case is therefore composed of three LeNet5, as described in Table\ref{mMNIST_archi}. The "Ensemble 3 classifiers" method also uses three LeNet5, while other methods are using two LeNet5, one for each modality.
We use dropout (50\% dropping) on the fully connected layers and batch normalization. The learning rate is 0.01, the batch size is 128 and the model is trained on 100 epochs for all experiments, except for Moddrop and Gated Multimodal Units, where hyper-parameters are found by a random grid search. Thus for Moddrop, the learning rate is changed into 0.05 and the modality drop probability is of 0.2. For Gated Multimodal Units, the dropout is changed into 25\% dropping.

First, we evaluate different alternatives for fusion (see Figure~\ref{schemaCentral}(b)) using element-wise sum, subtract and product, for several configurations of our toy dataset. The energy is in \{0.1, 0.25, 0.5\} and share ratio in \{0, 0.1, 0.5, 0.9\}, allowing to assess the improvement given by fusion on each configuration. We also evaluate the Fusion+Ensemble method, \textit{i.e.}, an ensemble of classifiers build on the top of the outputs of the fusion method (each modality make a prediction, as well as the fusion method, giving an ensemble of 3 classifiers). Finally, we also report the results of our CentralNet approach.

We numbered the layers of LeNet5 from 0 (input level) to 4 (prediction level) and evaluate the methods for the 5 different fusion depth, in order to find out which one yields is the best. Figure~\ref{energy} reports the performance of the different methods. The performance of the Fusion and Fusion+Ensemble methods are given in the case of their best fusion depth. 

These results first underline the proportionality relation between the energy per modality and the error rate. It is also worth noting that not sharing enough or too much information between the modalities lowers the accuracy and the interest of a fusion approach. This observation is in line with~\cite{baltruvsaitis2018multimodal,atrey2010multimodal}. 

As shown in Table~\ref{mMNIST_results} the optimal fusion layer obtained for each method differs but is early. Other properties are highlighted: A complementarity between Fusion and Ensemble exists, as shown by the improvement brought by the Fusion+Ensemble method. Nevertheless, as soon as the modalities share a large amount of information, the Ensemble method outperforms the Fusion method. It implies that the benefit of the fusion depends on the nature of the dataset and can be null.

Independently to the chosen configuration, our CentralNet approach achieves the best results, except in the case of a null share ratio (first point of the right-hand side of the Figure~\ref{energy}). In this case, the modalities are not sharing information, so the better performance of the Fusion+Ensemble (fusing at layer 0) compared to CentralNet might be explained by the difficulty to find relation between independent modalities and thus constructing a stable joint representation from the learned weighted sum.
A comparison with an Ensemble of 3 models  applied on original images suggests that this performance does not come only from a larger number of parameters.

Table~\ref{all_weights} shows that in the lowest layers of CentralNet, the modalities are taken into account, while on last layers the weight of central previous hidden layer dominates. This is in line with our observations on the Fusion+Ensemble results.

\subsection{Audiovisual MNIST}
\begin{table}[tb]
\centering
\begin{tabular}{l|l|l|l|l|ll}
\hline
                                                                               & \multicolumn{2}{l|}{Image} & \multicolumn{2}{l|}{Central} & \multicolumn{2}{l}{Audio}                 \\ \hline
Part                                                                           & Type     & Output size     & Type       & Output size     & \multicolumn{1}{l|}{Type}  & Output size  \\ \hline
\multirow{3}{*}{\begin{tabular}[c]{@{}l@{}}Features\\ Extraction\end{tabular}} &          &                 &            &                 & \multicolumn{1}{l|}{Conv1} & 56 x 56 x 8  \\
                                                                               &          &                 &            &                 & \multicolumn{1}{l|}{Conv2} & 28 x 28 x 16 \\
                                                                               & Conv1    & 14 x 14 x 32    &            &                 & \multicolumn{1}{l|}{Conv3} & 14 x14 x 32  \\ \hline \hline
\multirow{3}{*}{Fusion}                                                        & Conv     & 7 x 7 x 64      & Conv       & 7 x 7 x 64      & \multicolumn{1}{l|}{Conv}  & 7 x 7 x 64   \\
                                                                               & Dense    & 1024            & Dense      & 1024            & \multicolumn{1}{l|}{Dense} & 1024         \\
                                                                               & Pred     & 10              & Pred       & 10              & \multicolumn{1}{l|}{Pred}  & 10           \\ \hline
\end{tabular}
\caption{The architecture of the CentralNet model on the avMNIST dataset.}
\label{avMNIST_size}
\end{table}

Audiovisual MNIST is a novel dataset we created by assembling visual and audio features. The first modality, disturbed image, is made of the 28x28 PCA-projected MNIST images, generated as explained in the previous section, with only 25\% of the energy, to better assess the benefits of the fusion method. The second modality, audio, is made of audio samples on which we have computed 112x112 spectrograms. The audio samples are the pronounced digits of the Free Spoken Digits Database~\cite{fsdd} augmented by adding randomly chosen 'noise' samples from the ESC-50 dataset~\cite{piczak2015esc}, to reach the same number of examples as in MNIST (55000 training examples, 10000 testing examples).  

For processing the image modality, we use the LeNet5 architecture~\cite{lecun1998gradient}, as in the previous section. For the audio modality, we use a 6-layer CNN, adding two convolution-pooling blocks. The whole architecture is detailed on Table~\ref{avMNIST_size}. 

We use dropout (50\% dropping) on the fully connected layers and batch normalization. The learning rate is 0.001, the batch size is 128 and the model is trained on 100 epochs for all experiments, except Moddrop and Gated Multimodal Units, where hyper-parameters are found by a random grid search. Thus for Moddrop, the learning rate is changed into 0.005 and the modality drop probability is of 0.32. For Gated Multimodal Units, the dropout is changed into 35\% dropping.

\begin{table}[tb]
\centering
\begin{tabular}{ll}
\hline
Method                    & Accuracy       \\ \hline
Disturbed image           & 72.8  $\pm 0.3$         \\
Audio                     & 86.1   $\pm 0.15$        \\ \hline
Weighted mean             & 94.7  $\pm 0.12$         \\
Concat                    & 93.7   $\pm 0.17$       \\
Concat + Multi-Task         & {94.8  ${\pm 0.11}$} \\
Moddrop\cite{neverova2016moddrop}       & {94.8 ${\pm 0.10}$}       \\
Gated Multimodal Unit\cite{arevalo2017gated}  & 94.1   $\pm 0.14$        \\
\textbf{CentralNet} & \textbf{95.0  $\mathbf{\pm 0.12}$} \\ \hline
\end{tabular}

\caption{Accuracy on the audiovisual MNIST dataset.}
\label{avMNIST}
\end{table}
The performance is measured as the per sample accuracy on the 10000 test samples. We observe from Table~\ref{avMNIST}, that the fusion methods are all performing better than unimodal ones. Both ensembles, Moddrop and simple weighted mean yield good performance but CentralNet performs best. Figure~\ref{all_weights} shows that all the modalities are used at each layer, meaning that they all bring information.



\subsection{Montalbano}
The Montalbano dataset~\cite{escalera2014chalearn} gathers more than 14000 samples of 20 Italian sign gesture categories. These videos were recorded with a Kinect, capturing audio, skeleton joints, RGB and depth. The task is to recognize the gestures from the video data. The performance is measured as the macro accuracy, which is the average of the per class accuracy.
\begin{figure}[tb]
    \centering
    \includegraphics[width=\textwidth]{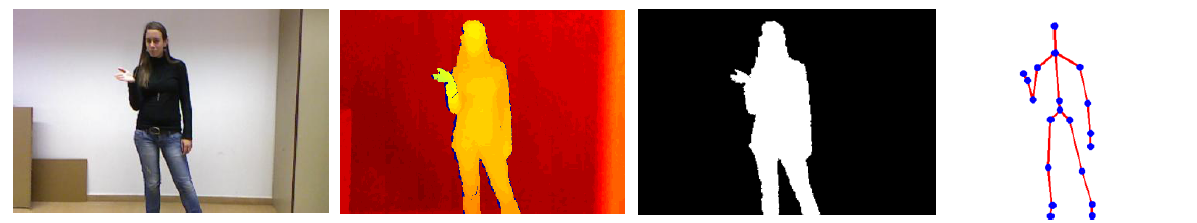}
    \caption{The different raw visual modalities provided by the organizers of the ChaLearn challenge on Montalbano dataset. Neverova \textit{et al.}~\cite{neverova2014multi} propose to focus on right and left hands, skeleton and audio.}
    \label{montalbano_moda}
\end{figure}

The features used in these experiments are those provided by Neverova \textit{et al.}~\cite{neverova2016moddrop}: audio features (size 350), motion capture of the skeleton (size 350), RGB+depth left/right hands features (size 400). Features are zero-padded (if needed) to give vectors of size 400. The fusion architecture includes one multilayer perceptron per modality, each having 3 layers of size: $400\times 128$,  $128\times 42$,  $42\times 21$. CentralNet architecture connects the 3 layers of the different modalities into a central network.

We use dropout (50\% dropping) and batch normalization. The learning rate is 0.05 (we multiply the learning rate by 0.96 at each epoch), the batch size is 42 containing two samples of each class and the model is trained on 100 epochs for all experiments. For Moddrop, the modality drop probability is of 0.5. 

\begin{table}[tb]
\centering
\begin{tabular}{l|l}
\hline
Method    & Accuracy \\ \hline
Left      & 46.0 $\pm 0.7$     \\
Audio     & 59.3 $\pm 0.3$     \\
Right     & 79.0 $\pm 0.3$     \\
Mocap     & 88.0 $\pm 0.3$     \\ \hline
Weighted mean    & 97.54 $\pm 0.02$    \\
Concat    & 97.76 $\pm 0.05$    \\
Concat + Multi-Task    & 98.02 $\pm 0.04$    \\
{Moddrop}   & {98.19 ${\pm 0.03}$}     \\
Gated Multimodal Unit &  97.98 $\pm 0.04$     \\
\textbf{CentralNet} & \textbf{98.27 $\mathbf{\pm 0.03}$}     \\ \hline

\end{tabular}

    \caption{Accuracy on the Montalbano validation set (same protocol as~\cite{neverova2016moddrop}).}
    \label{table-res-montalbano}
\end{table}
Table~\ref{table-res-montalbano} shows that the performance obtained with each modality varies from 46\% (left hand) to 88\% (mocap). Basic late fusion gives significant improvement, suggesting complementarity between modalities. CentralNet outperforms all other approaches. 
Figure~\ref{all_weights} shows the weights of the different modalities at each level. At the first layer (layer 0), the weights reflect the dimensionality of the layers. At the next layer, almost no information is taken from the modalities, while at layers 2 and 3, the weight given to each modality and to the central representation are relatively similar. This may be interpreted as an hybrid fusion strategy, mixing "early" and "late" fusions.

\subsection{MM-IMDb}
\begin{figure}[tb]
    \centering
    \includegraphics[width=0.98\textwidth]{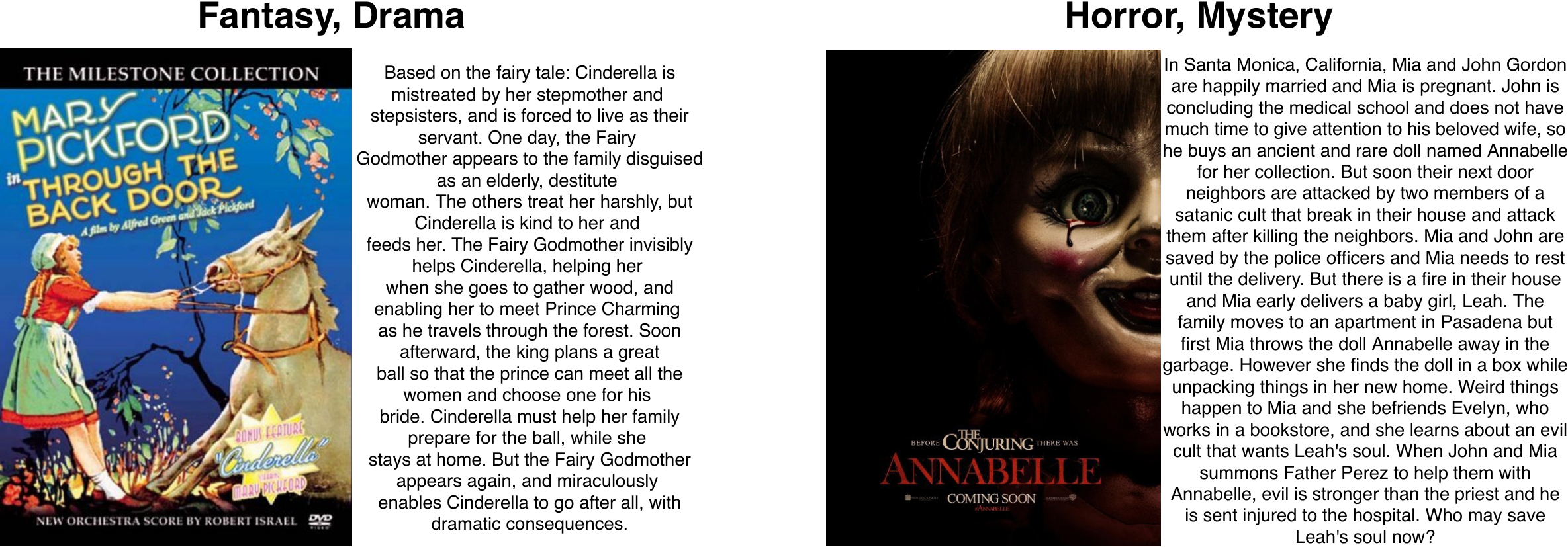}
    \caption{Two movie samples extracted from the mm-IMDB dataset. For each, we can see the poster and the associated plot. The genres to predict are displayed on the top of the figure.}
    \label{imdbposter}
\end{figure}
The MM-IMDb dataset~\cite{arevalo2017gated} comprises respectively 15552, 2608 and 7799 training, validation and test movies, along with their plot, poster, genres and other 50 additional metadata fields such as year, language, writer, director, aspect ratio, \textit{etc.}. The task is to predict a movie genre based on its plot and on its poster (\textit{cf.} Figure~\ref{imdbposter}). One movie can belong to more than one of the 23 possible genres. The task hence has to be evaluated as a multilabel classification task. As in~\cite{arevalo2017gated,kiela2018efficient}, we measure the performance with the micro, macro, weighted and per sample F1 scores.  
For these experiments, we use the features kindly provided by the authors~\cite{arevalo2017gated}. The visual feature of size 4096 is extracted from the posters using the VGG-16~\cite{simonyan2014very} network pretrained on Imagenet. The 300-d textual one are computed with a fine-tuned word2vec~\cite{mikolov2013distributed} encoder. 

We build a multilayer perceptron on the top of the features of each modality. For both modalities, the network has 3 layers of size $\textit{input$\_$size}\times 4096$, $4096\times 512$ and $512\times 23$. The CentralNet architecture (see Table~\ref{mmIMDB_archi} is the same, taking 4096-d vectors as inputs, zero-padding the textual features to reach the visual features size.
\begin{table}[tb]
\centering
\begin{tabular}{ll|l|l|l|l}
\hline
\multicolumn{2}{l}{Text}          & \multicolumn{2}{l|}{Central} & \multicolumn{2}{l}{Visual} \\ \hline
\multicolumn{1}{l|}{Type}  & Size & Type          & Size         & Type          & Size        \\ \hline
\multicolumn{1}{l|}{Dense} & 2048 & Dense         & 2048         & Dense         & 2048        \\
\multicolumn{1}{l|}{Dense} & 512  & Dense         & 512          & Dense         & 512         \\
\multicolumn{1}{l|}{Pred}  & 23   & Pred          & 23           & Pred          & 23          \\ \hline
\end{tabular}
\caption{Architecture of the CentralNet on the MM-IMDb dataset.}
\label{mmIMDB_archi}
\end{table}

We use dropout (50\% dropping) and batch normalization. The learning rate is 0.01 and the batch size is 128. For Moddrop, the modality drop probability is of 0.25. The loss of the models is a cross entropy, but we put a weight of 2.0 on the positives terms to balance precision and recall. More formally, the loss is:
\begin{equation}
    loss = -\log(2\sigma(pred))y - (1 - y) log(1 - \sigma(pred))
\end{equation}
with $\sigma(pred)$ the sigmoid activation of the last output of the network and $y$ the multiclass label.
As recommended by Arevalo~\textit{et al.}~\cite{arevalo2017gated}, we also use early stopping on the validation set.

Table~\ref{table_imdb_results} reports the performance measured during the different experiments. First of all, the worst confidence interval we observe is very small, of the order of $\pm 0.001$. For making the table more readable, we do not include it.  
\begin{table}[tb]
\centering
\begin{tabular}{lllll}
\hline
Method                           & Micro          & Macro          & Weighted       & Samples        \\ \hline
Text (alone)                            & 0.602          & 0.489          & 0.585          & 0.606          \\
Image  (alone)                           & 0.478          & 0.256          & 0.421          & 0.484          \\ \hline

Weighted mean                    & 0.635          & 0.550          & 0.626          & 0.634          \\
Concat                           & 0.611          & 0.506          & 0.599          & 0.614          \\
Concat + Multi-Task                & 0.623          & 0.528          & 0.613          & 0.622          \\
Moddrop\cite{neverova2016moddrop}            & 0.624          & 0.526          & 0.614          & 0.625           \\
{Gated Multimodal Unit~\cite{arevalo2017gated}}                     & {0.630} & {0.541} & {0.617} & {0.630} \\
\textbf{CentralNet} & \textbf{0.639} & \textbf{0.561} & \textbf{0.631} & \textbf{0.639} \\ \hline
\end{tabular}
\caption{F1 scores of the different methods on the MM-IMDb test set.}
\label{table_imdb_results}

\end{table}
Second, one can observe that the textual modality clearly outperforms the visual one. Third, we note that even the basic fusion methods, such as the concatenation of the features, improve the score. Finally, the Concat+Multi-Task and Concat+ModDrop methods are outperformed by a significant margin by Gated Multimodal Unit and CentralNet, which is giving the best performance. Figure~\ref{all_weights} shows that CentralNet gives more weight to the first layers, indicating that an "early fusion" strategy is privileged in this case, even if the two modalities contribute significantly at all levels. 
\section{Conclusions}
This paper introduced a novel approach for the fusion of multimedia information. It consists in a joint representation having the form of a central network connecting the different layers of modality specific neural networks. The loss of this central network not only allows to learn how to combine the different modalities but also adds some constraints on the modality specific networks, enforcing their complementary aspects. This novel model achieves state-of-the-art results on several different multimodal problems. It also addresses elegantly the {\em late versus early} fusion paradigm.

\bibliographystyle{splncs}
\bibliography{egbib}
\end{document}